\documentclass{article}

\usepackage{arxiv}

\usepackage[utf8]{inputenc} 
\usepackage[T1]{fontenc}    
\usepackage{hyperref}       
\usepackage{url}            
\usepackage{booktabs}       
\usepackage{amsfonts}       
\usepackage{nicefrac}       
\usepackage{microtype}      
\usepackage{lipsum}
\usepackage{graphicx}
\graphicspath{ {./images/} }

\usepackage{amsmath}
\usepackage{amssymb}

\usepackage{booktabs}

\usepackage{amsmath,amsfonts}
\usepackage{array}
\usepackage[caption=false,font=normalsize,labelfont=sf,textfont=sf]{subfig}
\usepackage{textcomp}
\usepackage{stfloats}
\usepackage{url}
\usepackage{verbatim}
\usepackage{graphicx}
\usepackage{cite}
\usepackage{hyperref}
\usepackage{url}
\usepackage{wrapfig}
\usepackage{graphicx}
\usepackage{amsmath}
\usepackage{amsthm}
\usepackage{amssymb}
\usepackage{booktabs}
\usepackage[switch]{lineno}
\usepackage{multirow}
\usepackage{cleveref}
\usepackage[ruled,vlined]{algorithm2e} 
\usepackage{float}
\usepackage{placeins}
\usepackage{adjustbox}

\usepackage{tikz}
\usetikzlibrary{positioning,calc,fit,decorations.pathmorphing,shapes.geometric,shapes.gates.logic.US,calc}
\usepackage{tikz-3dplot}

\usepackage{pgfplots}
\usepackage{pgfplotstable}
\usepgfplotslibrary{groupplots}
\usepackage{wrapfig}
\usepackage{xcolor}
\usetikzlibrary{arrows.meta,positioning,fit,shadows}

\usepackage{threeparttable}
\usepackage{array}
\usepackage{graphicx}
\usepackage{wrapfig} 
\usepackage{amsmath}
\usepackage{amssymb}

\usepackage{algpseudocode}

\usepackage[utf8]{inputenc}
\usepackage[T1]{fontenc}
\usepackage{geometry}
\usepackage{times}
\usepackage{graphicx}
\usepackage{caption}
\usepackage{lipsum} 

\usepackage{tikz}
\usepackage{pgfplots}
\pgfplotsset{compat=1.18}
\usetikzlibrary{patterns}

\usepackage{tikz}
\usetikzlibrary{arrows.meta, positioning}
\usepackage[dvipsnames,table,xcdraw]{xcolor}

\title{Towards Efficient Agents: A Co-Design of Inference Architecture and System}

\author{
 Weizhe Lin, Hui-Ling Zhen, Shuai Yang, Xian Wang, Renxi Liu, Hanting Chen, \\ 
  \textbf{ Wangze Zhang, Chuansai Zhou, Yiming Li, Chen Chen, Xing Li, Zhiyuan Yang,} \\ 
 \textbf{Xiaosong Li, Xianzhi Yu, Zhenhua Dong, Mingxuan Yuan, Yunhe Wang}\\\\ 
\textbf{ Huawei}
}

\begin{document}
\maketitle
\begin{abstract}
The rapid development of large language model (LLM)--based agents
has unlocked new possibilities for autonomous multi-turn reasoning and tool-augmented decision-making. However, their real-world deployment is hindered by severe inefficiencies that arise not from isolated model inference, but from the systemic latency accumulated across reasoning loops, context growth, and heterogeneous tool interactions. This paper presents AgentInfer, a unified framework for end-to-end agent acceleration that bridges inference optimization and architectural design. We decompose the problem into four synergistic components: AgentCollab, a hierarchical dual-model reasoning framework that balances large- and small-model usage through dynamic role assignment; AgentSched, a cache-aware hybrid scheduler that minimizes latency under heterogeneous request patterns; AgentSAM, a suffix-automaton–based speculative decoding method that reuses multi-session semantic memory to achieve low-overhead inference acceleration; and AgentCompress, a semantic compression mechanism that asynchronously distills and reorganizes agent memory without disrupting ongoing reasoning. Together, these modules form a Self-Evolution Engine capable of sustaining efficiency and cognitive stability throughout long-horizon reasoning tasks. Experiments on the BrowseComp-zh and DeepDiver benchmarks demonstrate that through the synergistic collaboration of these methods, AgentInfer reduces ineffective token consumption by over 50\%, achieving an overall 1.8×-2.5× speedup with preserved accuracy. These results underscore that optimizing for agentic task completion—rather than merely per-token throughput—is the key to building scalable, efficient, and self-improving intelligent systems.

\end{abstract}

\section{Introduction}

Autonomous agents~\cite{chen2025pangu,huang2025deep}, powered by large language models (LLMs), are poised to revolutionize domains ranging from deep scientific research and iterative software engineering to knowledge-intensive decision-making. Unlike conventional single-turn LLM applications—where the model generates a one-off response and terminates—autonomous agents operate within a persistent \textit{Think--Act--Observe} loop. In this cycle, the agent formulates plans, interacts with external tools (e.g., search engines, simulators, code interpreters), observes outcomes, and iteratively reasons based on accumulated context. While this continuous, reflection-driven process endows agents with remarkable problem-solving flexibility, it imposes severe latency penalties and computational overheads. Each ``Think'' step typically triggers a large model inference involving extended contexts, parallel requests, and complex reasoning paths, resulting in cumulative delays that hinder real-time deployment and scalability.

A fundamental dichotomy exists between agent acceleration and traditional LLM inference optimization. Conventional speedups—such as quantization~\cite{li2025kvtuner}, pruning~\cite{chen2025pangu,pei2025cmoe,pei2024fusegpt}, or speculative decoding~\cite{sun2025scaling}—predominantly focus on \textit{intra-step} efficiency, aiming to maximize tokens per second (TPS). In contrast, the efficiency of an autonomous agent is defined by the \textit{inter-step} time-to-completion of multi-turn tasks. A marginal gain in decoding speed is negligible if the agent, due to degraded reasoning quality, requires dozens of additional correction cycles. Thus, optimizing an agent system demands a paradigm shift: bridging intra-step acceleration (faster inference) with inter-step efficacy (fewer steps to solution)~\cite{wu2025resum,lin2025trimr}.

Consequently, traditional metrics like TPS can be perilously misleading in agentic contexts. An agent is not merely a passive text generator but an active reasoning system where progress is path-dependent~\cite{java2025characterizing}. A ``faster but less accurate'' model often regresses overall performance by introducing factual errors, invalid tool calls, or incoherent summaries that cascade through subsequent turns. Such errors necessitate retries and redundant inference, negating any low-level speedups. Therefore, agent performance must be evaluated through holistic metrics that consider \textit{End-to-End Task Latency} (wall-clock time), \textit{Total Computational Cost} (aggregate FLOPs), and \textit{Turn Efficiency} (reasoning stability).

This disconnect stems from the cyclic and state-dependent nature of agent workflows. Unlike static one-shot inference, agent reasoning unfolds dynamically—each thought is conditioned on evolving memory and environmental feedback. Optimization strategies that treat inference calls in isolation fail to exploit cross-step dependencies, context reuse, and adaptive precision—opportunities unique to agentic workloads. To achieve system-level efficiency, acceleration techniques must be designed to optimize the entire reasoning trajectory rather than isolated invocations.

In this report, we focus primarily on \textbf{Deep Research Agents}—an emerging class of autonomous systems designed for multi-round reasoning, tool-augmented exploration, and evidence-based synthesis. Due to their iterative reasoning patterns, heavy reliance on contextual memory, and extensive tool interactions, these agents serve as a representative and rigorous testbed for studying end-to-end agent acceleration.

We argue that accelerating agent inference requires a \textbf{synergistic, multi-layered framework} spanning both the reasoning process and the inference engine. To this end, we introduce \textbf{AgentInfer}, a hierarchical framework that systematically enhances agent efficiency. While our principles apply to general agentic systems (e.g., coding or dialogue agents), we validate them within the demanding context of Deep Research Agents. 
At the reasoning level, we propose \textbf{\textit{AgentCollab}}, a self-evaluation-driven dual-model mechanism. It delegates routine inference and tool execution to a smaller model, while a larger model handles strategic planning and rescues stalled trajectories. This is governed by structured \textit{Progress Check} signals, which trigger escalation only when necessary, thereby conserving expensive capacity. Complementarily, \textbf{\textit{AgentCompress}} implements lightweight context summarization, pruning redundant reasoning traces to maintain compact yet semantically complete histories.
At the inference-engine level, \textbf{\textit{AgentSAM}} advances speculative decoding~\cite{hu2024samdecodingspeculativedecoding} by using a suffix automaton to efficiently reuse token-level predictions across agent sessions.
Simultaneously, \textbf{\textit{AgentSched}} introduces a hybrid, KVCache-aware scheduling policy. By utilizing a dynamic shadow-price controller, it adaptively switches between Shortest-Job-First and cache-preserving modes, balancing latency and cache hit rates under dynamic workloads. Together, these components form a closed-loop architecture that integrates cognitive delegation, speculative decoding, adaptive scheduling, and context compression—optimizing the reasoning lifecycle in its entirety.

The primary contributions of this technical report are:
\begin{itemize}
    \item We propose \textbf{AgentInfer}, a systematic framework unifying reasoning-level and system-level optimizations. We demonstrate that achieving true efficiency in autonomous agents requires a top-down, multi-layered paradigm rather than isolated speedups.
    
    \item AgentCollab, AgentSAM, AgentSched, and AgentCompress are four components, and each incorporating customized acceleration strategies tailored to the unique characteristics of agent workloads.
    
    \item We provide comprehensive empirical evidence showing that integrating these multi-level techniques yields substantial improvements in end-to-end latency and reliability across complex agent benchmarks.
\end{itemize}

\section{Related Work}
\label{sec:related_work}

Large language models (LLMs) often form the reasoning core of agent systems, but their computational and memory demands limit interactive efficiency. 
Techniques such as model quantization, pruning, and low-rank adaptation have been widely investigated to reduce inference latency. 
Post-training quantization methods (e.g., W8A8, GPTQ, SmoothQuant) have achieved tangible speedups but may degrade reasoning accuracy on multi-step or symbolic tasks~\cite{dettmers2023qlora,frantar2023gptq}. 
Recent works like QLoRA~\cite{dettmers2023qlora} and AWQ~\cite{lin2024awq} aim to maintain performance via adaptive weight grouping and normalization-aware tuning. 
In addition, distillation-based models~\cite{hinton2015distilling,wang2024tinyllm} compress high-capability agents into lightweight student models for low-latency scenarios. 
These methods focus primarily on per-token acceleration but do not directly address multi-turn agent bottlenecks such as context growth or reasoning instability.

Context explosion is a major concern in agent execution loops, as the history of reasoning, tool calls, and observations grows over time.
To handle this, retrieval-augmented generation (RAG)~\cite{lewis2020rag,gao2023retrieval} integrates external memory to avoid long prompt reuse. 
Hierarchical summarization and ephemeral memory~\cite{xu2024memorybank,chen2024retentive} further compress intermediate interactions while preserving key task-specific cues. 
However, summarized or lossy memories can introduce \textit{semantic drift} between an agent’s reasoning state and its retrieved context, a phenomenon recently observed in multi-turn setups~\cite{park2023generativeagents}. 
More persistent memory mechanisms—such as vector-space long-term memory stores or semantic caches—are being explored to ensure contextual continuity across sessions~\cite{yuan2024longcontext,liu2024memgpt}.

Reasoning efficiency can be improved by both algorithmic and architectural advances. 
Tree-of-Thoughts (ToT)~\cite{yao2024treeofthoughts} and Reflection Loop agents~\cite{shinn2023reflexion} structure inference into tree or iterative reasoning loops, trading additional calls for stability and correctness. 
To mitigate the resulting latency, hybrid strategies such as speculative decoding~\cite{leviathan2023fast,sun2025scalingupspeedingup}, early-exit reasoning~\cite{zhang2024eegpt}, and dynamic reasoning depth~\cite{xie2024dynapipe} have been introduced. 
These frameworks parallelize or truncate reasoning trajectories while keeping semantic fidelity, yielding near-optimal balance between compute and accuracy.

At the system level, multi-agent orchestration frameworks~\cite{schick2023toolformer,gao2024autogen} expose severe heterogeneity in request length and tool usage patterns. 
Efficient scheduling algorithms—such as cache-aware dispatching~\cite{liu2025lmcache}, hybrid batching~\cite{yang2025kvshare}, or pipelined execution~\cite{zhang2025clusterattn}—have shown significant throughput gains for heterogeneous workloads. 
Moreover, persistent KV-cache management~\cite{haoyang2025survey,li2025kvtuner} and prefix-sharing mechanisms improve utilization for long interactions, mitigating recomputation costs during prefill stages. 
Recent open systems like \textit{vLLM}~\cite{vllm} and \textit{SGLang}~\cite{sglang} provide foundational infrastructures supporting efficient agent orchestration across distributed environments.

Overall, while numerous acceleration strategies exist for LLMs and dialog systems, relatively few directly address the \emph{emergent inefficiencies of reasoning-driven agents}. 
An efficient agent system must consider not only per-token speed but also multi-turn execution stability, context alignment, and cache-aware scheduling—dimensions where existing methods remain fragmented. 
Our work aims to unify these perspectives by analyzing and optimizing the full agent execution pipeline end-to-end.
\section{Agent Workflows}

The operational core of a reasoning-based or multi-turn agent is its \textit{execution loop}—a recurrent process that interleaves reasoning, actuation, and perception. In each iteration, the agent engages in a structured cycle:

\begin{enumerate}
    \item \textbf{Think:} The agent interprets its current goal and contextual history, performs internal reasoning, and formulates a plan or hypothesis for the next step. This stage typically corresponds to one or more inference calls to a large language model (LLM).
    \item \textbf{Act:} Based on the reasoning outcome, the agent invokes external tools or APIs to obtain new information or execute actions (e.g., \texttt{web\_search("...")}, \texttt{python("code")}, or knowledge-base retrieval).
    \item \textbf{Observe:} The agent processes and integrates the tool output into its evolving context, enabling subsequent reasoning grounded on updated environmental state.
\end{enumerate}

In certain multi-agent frameworks featuring explicit agent orchestration (e.g., \textbf{DeepDiverV2}~\cite{openpangu_deepdiver_v2}), the central controller is referred to as the \textbf{Planner Agent}.
It is responsible for decomposing complex objectives into well-defined sub-tasks and coordinating the execution flow among subordinate agents.
Once the overall task plan is established, the Planner Agent dynamically instantiates and supervises specialized sub-agents to perform targeted operations—such as the \textbf{Information Seeker Agent}, which conducts knowledge retrieval and evidence collection, and the \textbf{Writer Agent}, which integrates retrieved information, synthesizes insights, and produces the final structured output.
This hierarchical orchestration design enables collaborative reasoning and division of labor among heterogeneous agents, thereby improving both the efficiency and robustness of the overall problem-solving process.

This iterative loop continues until a termination condition (e.g., goal achievement or user-defined limit) is reached. While conceptually simple, such a pattern introduces two fundamental bottlenecks that dominate runtime performance:

\begin{itemize}
    \item \textbf{Inference Latency:} Each \textit{Think} phase triggers a synchronous LLM inference, where token-by-token autoregressive decoding incurs tens to hundreds of milliseconds per token. As the number of steps grows, accumulated inference delay becomes the dominant contributor to wall-clock latency. This is particularly exacerbated in multi-agent setups, where concurrent reasoning calls amplify resource contention.

    \item \textbf{Context Explosion:} Every new iteration appends reasoning traces, tool invocations, and observations to the agent’s prompt. The prompt length therefore, increases approximately linearly with the number of turns, but the computational cost of self-attention grows quadratically with sequence length. Without summarization or memory management mechanisms, this leads to rapidly increasing latency, degraded attention precision, and eventual truncation beyond the model’s context window.
\end{itemize}

\section{The Efficiency Paradox: Why Standard Optimizations Fail Agents}

While standard optimization techniques—such as post-training quantization, context summarization, and continuous batching—have proven highly effective for stateless, single-turn LLM applications, they often yield counterproductive results when applied to autonomous agents. This section analyzes three critical dimensions where these conventional methods fail to align with the stateful, path-dependent nature of agentic workloads.

\subsection{The Quantization Trap} 

In deep-research agentic workflows, a fundamental conflict exists between inference throughput and reasoning fidelity. While low-bit quantization (e.g., W8A8) is standard for reducing memory bandwidth in chat applications, our analysis reveals it is detrimental to agents requiring strict syntactic precision.

As detailed in Table~\ref{tab:quantization_trap}, blindly applying quantization creates a \textit{``Latency Trap''}. Although INT8 quantization accelerates atomic forward passes by 45\%, it degrades the agent's ability to perform precise operations—such as mathematical derivation and JSON formatting. This precision loss triggers a cascade of failures, forcing the agent into expensive self-correction loops. Consequently, while the \textit{cost per token} decreases, the \textit{total time-to-solution} increases by 70\% due to the substantial overhead of retries.

\begin{table}[htbp]
\centering
\caption{\textbf{The Quantization Trap.} Comparison of FP16 baseline versus INT8 quantization on the HLE benchmark. Although quantization improves throughput (TPS), the loss in precision causes a sharp decline in success rate and triggers a 3.5$\times$ increase in retries, ultimately inflating the total time-to-solution.}
\label{tab:quantization_trap}
\resizebox{0.95\linewidth}{!}{%
\begin{tabular}{lcccc}
\toprule
\textbf{Metric} & \textbf{FP16 (Baseline)} & \textbf{INT8 (Quantized)} & \textbf{$\Delta$ Relative} & \textbf{Impact} \\ 
\midrule
\textbf{Throughput} (Tokens/s) & 42.5 & 61.6 & \textcolor{blue}{+45.0\%} & Faster Inference \\
\textbf{Task Success Rate} & 88.2\% & 61.7\% & \textcolor{red}{-30.0\%} & Degradation \\
\midrule
\textit{Breakdown of Latency:} & & & & \\
\quad Avg. Inference Time (s) & 45.0 & 30.0 & \textcolor{blue}{-33.3\%} & Step Speedup \\
\quad Avg. Recovery Time (s) & 5.0 & 55.0 & \textcolor{red}{+1000\%} & \textbf{Retry Loop} \\
\textbf{Total Time-to-Solution} (s) & \textbf{50.0} & \textbf{85.0} & \textcolor{red}{+70.0\%} & \textbf{Slower End-to-End} \\
\bottomrule
\end{tabular}%
}
\end{table}

\subsection{Granularity Mismatch}

To mitigate the quadratic cost of attention in long-context scenarios, summarization is often employed to compress context. However, we identify a critical \textit{granularity mismatch} for agents. Deep research agents necessitate high-fidelity access to raw details—such as specific variable names in code snippets or numerical values in financial reports—to execute tools correctly.

Table~\ref{tab:summarization_gap} illustrates this inverse correlation. Summarization acts as a lossy compression layer, stripping away syntactic details deemed ``redundant'' by the proxy model. This ambiguity forces the agent to issue multiple clarification queries to recover missing information. As a result, the reduction in tokens per step is negated by a 3.5$\times$ increase in interaction turns, leading to comparable total token consumption but a significantly higher risk of context drift.

\begin{table}[htbp]
\centering
\caption{\textbf{The Summarization Gap.} While summarization reduces the context window size per step, it introduces ambiguity that forces the agent to perform more turns to clarify information. This ``Scissor Effect'' neutralizes the efficiency gains.}
\label{tab:summarization_gap}
\resizebox{0.75\linewidth}{!}{%
\begin{tabular}{lccc}
\toprule
\textbf{Metric} & \textbf{Full Context} & \textbf{Summarized} & \textbf{Observation} \\ 
\midrule
\textbf{Avg. Tokens per Step} & 8,500 & 2,100 & \textcolor{blue}{4$\times$ Reduction} \\
\textbf{Avg. Turns to Solve} & 4.0 & 14.0 & \textcolor{red}{3.5$\times$ Increase} \\
\midrule
\textbf{Total Token Consumption} & $\approx$ 34k & $\approx$ 29.4k & Marginal Gain \\
\textbf{Context Drift Rate} & Low & High & \textbf{Cognitive Ambiguity} \\
\bottomrule
\end{tabular}%
}
\end{table}

\subsection{The Memory Persistence Bottleneck}

Modern LLM serving engines typically utilize Shortest Job First (SJF) scheduling, optimizing for stateless, high-throughput chat workloads. This design is fundamentally misaligned with agentic workloads, which are characterized by long-context, multi-turn sessions requiring \textit{memory persistence}.

As shown in Table~\ref{tab:scheduling_impact}, standard schedulers prioritize short incoming requests, aggressively evicting the KV-cache of idle agents waiting for tool execution. Upon resumption, the system must re-compute the entire historical context (Prefill). For agents with contexts exceeding 32K tokens, this leads to a cache hit rate of only 15\%, inducing massive latency spikes (up to 3.1s). This demonstrates that agent efficiency relies not merely on compute speed, but on the \textit{temporal locality} of memory management.

\begin{table}[htbp]
\centering
\caption{\textbf{Impact of Scheduling Policy on Memory Persistence.} Standard SJF scheduling favors short-context agents and causes frequent KV-cache eviction for long-context agents, leading to low cache hit rates and high prefill latency.}
\label{tab:scheduling_impact}
\resizebox{0.7\linewidth}{!}{%
\begin{tabular}{lcc}
\toprule
\multirow{2}{*}{\textbf{Workload Type}} & \multicolumn{2}{c}{\textbf{Standard SJF}} \\
\cmidrule(lr){2-3}
& \textbf{Hit Rate} & \textbf{Prefill Latency} \\ 
\midrule
Short Context ($<$4K) & 92\% & 150 ms \\
\textbf{Long Context ($>$32K)} & \textbf{15\%} & \textbf{3,100 ms} \\
\midrule
\textit{Observation} & \multicolumn{2}{c}{\textit{High eviction and re-computation for long contexts}} \\
\bottomrule
\end{tabular}%
}
\end{table}

\section{AgentInfer: Framework for Efficient Agent} 

\begin{figure*}[htbp]
    \centering
    \includegraphics[%
        width=0.94\linewidth,%
        clip,%
        trim=0 1.5cm 0 0
    ]{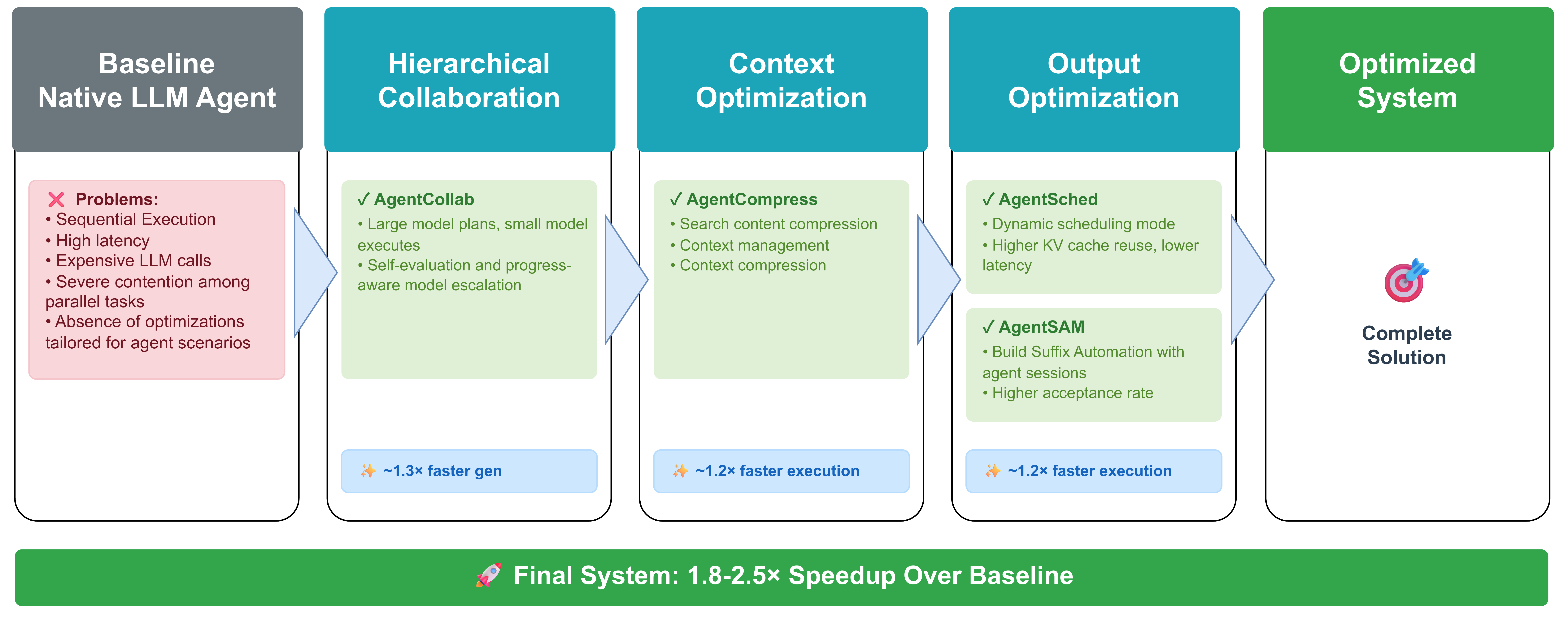}
    \caption{Overview of modules in AgentInfer.}
    \label{fig:agent_infer_framework}
\end{figure*}

In this technical report, we propose a progressive optimization framework for LLM agents, as illustrated in Figure~\ref{fig:agent_infer_framework}. The framework decomposes the end-to-end reasoning and execution pathway into a sequence of cooperative acceleration modules, each designed to mitigate a specific latency or inefficiency source across the agent lifecycle. The pipeline begins with a naïve baseline agent that performs sequential reasoning and synchronous tool invocations, resulting in high latency and cost.

\textbf{Stage~1 (Difficulty-Aware Collaboration)} introduces \textit{AgentCollab}, a dual-model collaboration mechanism in which a larger model performs a high-level planning and problem decomposition, and a smaller model then takes over most routine reasoning and tool calls; whenever the searching trajectory is stagnating, the session is temporarily escalated to the large model until progress resumes, thereby concentrating expensive large-model usage on genuinely hard segments while keeping overall inference cost under control.

\textbf{Stage~2 (Context Optimization)} employs \textit{AgentCompress}, a lightweight self-compression and filtering layer that prunes invalid or redundant search results, compresses context, reduces context length, and enhances overall compute efficiency. 

\textbf{Stage~3 (Execution Acceleration)} incorporates two complementary modules: \textit{AgentSAM}, an improved speculative decoding algorithm that accelerates token generation using cross-session agent memory; and \textit{AgentSched}, a hybrid, KV-aware scheduling mechanism that dynamically balances short-job responsiveness and cache reuse efficiency. Together, these modules enhance real-time concurrency while minimizing context eviction and redundant recomputation.

Finally, the optimized system forms a closed-loop design that hierarchically integrates cognitive guidance, predictive token speculation, dynamic scheduling, and adaptive compression. 
Compared with conventional LLM acceleration techniques that focus primarily on isolated kernel or quantization improvements, our framework introduces a hierarchy of optimizations tailored specifically for agentic workloads. 
Each module contributes complementary gains under different operational conditions. 
These enhancements accumulate in a compositional manner, yielding consistent, scalable benefits across agentic workloads, and ultimately achieving more than $50\%$ token usage reduction and $1.8$--$2.5\times$ end-to-end latency acceleration.

\subsection{AgentCollab: Self-Evaluation-Driven Multi-Agent Collaboration}

\subsubsection{Methodology} 

Deep-research-style agents require both strong reasoning ability and strict control over inference cost. To reconcile these goals, we propose \textit{AgentCollab}, a self-evaluation-driven collaboration framework that coordinates a large model and a small model through dynamic escalation and de-escalation.

The large model ($M_L$) acts as a high-capability reasoner responsible for initial task framing and for rescuing trajectories that become stuck. At the beginning of a session, $M_L$ executes a small number of `Think' steps to: (1) assess user intent and task difficulty, (2) outline a high-level solution strategy, and (3) decompose the problem into intermediate objectives and tool-usage plans. After this warm-up phase, the controller switches to the small model ($M_S$), which executes the majority of subsequent `Think' steps and tool calls, benefiting from much lower per-token cost.

At the core of AgentCollab is a \textbf{self-evaluation subroutine}, instantiated as a structured \emph{Progress Check} block. After each `Think' step, the currently active model is required to emit a \texttt{PROGRESS} block in the following canonical format:
\begin{verbatim}
===PROGRESS===
<reason> ...rationale... </reason>
<value> TRUE or FALSE </value>
===END_PROGRESS===
\end{verbatim}
Here, \texttt{<reason>} provides a concise meta-cognitive justification of whether the latest actions have meaningfully advanced the trajectory toward the final goal, and \texttt{<value>} encodes a binary judgment of \emph{significant progress}. When the active model is $M_S$ and \texttt{<value>} is \texttt{TRUE}, the controller continues using the small model; simple or well-structured tasks are therefore handled almost entirely by $M_S$. When \texttt{<value>} is \texttt{FALSE}, this self-diagnostic signal is interpreted as a sign of difficulty (e.g., oscillation, redundant probing, or failure to unlock new information), and the session is \emph{escalated} to $M_L$.

During escalation, $M_L$ takes over subsequent `Think' steps, re-analyzing the current context, correcting potential reasoning errors, and—if necessary—reframing the plan. The same self-evaluation protocol is applied to the large model's `Think' steps: each step yields a Progress Check block that assesses whether the trajectory has exited the “stuck” region. Once $M_L$ produces one or more `Think' steps whose Progress Check returns \texttt{TRUE}, indicating that the trajectory has resumed meaningful progress, the controller \emph{de-escalates} back to $M_S$. This cycle may repeat multiple times within a single trajectory.

This self-evaluation-driven collaboration paradigm preserves the efficiency of small models on easy or moderately difficult segments while selectively leveraging the semantic robustness of large models only when needed. Difficulty-aware escalation prevents the agent from wasting tokens on fruitless local search, while avoiding the overhead of invoking the large model at every reasoning step.

\subsubsection{Algorithm Design}  

We now formalize the AgentCollab controller as a \textit{Self-Evaluation-Driven Escalation Architecture}, summarized in Algorithm~\ref{alg:agentcollab}. The core insight is that multi-step reasoning can be viewed as a sequence of `Think'--`Evaluate' cycles, where each cycle is assessed for \emph{incremental progress} rather than only final correctness. AgentCollab exploits this trajectory-level self-evaluation to decide when to switch between $M_L$ and $M_S$.

\begin{algorithm}[H]
\caption{AgentCollab with Self-Evaluation-Driven Escalation}
\label{alg:agentcollab}
\small
\KwIn{User query $Q$, Large model $M_L$, Small model $M_S$, Initial large-model budget $K_L$, Maximum consecutive large-model steps per escalation $B_L$}
\KwOut{Final answer $A$}

\tcp{Phase 0: Initialization}
$ctx \gets \text{init\_context}(Q)$\;
$mode \gets \text{LARGE}$\;
$large\_steps\_used \gets 0$\;

\tcp{Phase 1: Initial Planning with Large Model}
\While{$large\_steps\_used < K_L$}{
    $think\_out \gets M_L.\text{THINK}(ctx)$\;
    $ctx \gets ctx \cup \{think\_out\}$\;
    $prog \gets M_L.\text{PROGRESS\_CHECK}(ctx)$\;
    $large\_steps\_used \gets large\_steps\_used + 1$\;
    \If{$prog.value = \text{TRUE}$}{
        \tcp{Early exit if task is already solved}
        \If{$\text{is\_final\_answer}(ctx)$}{
            $A \gets \text{extract\_answer}(ctx)$\;
            \Return{$A$}\;
        }
    }
}
$mode \gets \text{SMALL}$\;

\tcp{Phase 2: Collaborative Reasoning with Self-Evaluation-Driven Escalation}
\While{\textbf{not} $\text{is\_final\_answer}(ctx)$}{
    \uIf{$mode = \text{SMALL}$}{
        $think\_out \gets M_S.\text{THINK\_AND\_TOOLS}(ctx)$\;
        $ctx \gets ctx \cup \{think\_out\}$\;
        $prog \gets M_S.\text{PROGRESS\_CHECK}(ctx)$\;
        
        \uIf{$prog.value = \text{TRUE}$}{
            \tcp{Small model is making progress; keep using it}
            \textbf{continue}\;
        }
        \Else{
            \tcp{Self-evaluation indicates stagnation; escalate to large model}
            $mode \gets \text{LARGE}$\;
            $large\_steps\_used \gets 0$\;
            \textbf{continue}\;
        }
    }
    \ElseIf{$mode = \text{LARGE}$}{
        $think\_out \gets M_L.\text{THINK\_AND\_TOOLS}(ctx)$\;
        $ctx \gets ctx \cup \{think\_out\}$\;
        $prog \gets M_L.\text{PROGRESS\_CHECK}(ctx)$\;
        $large\_steps\_used \gets large\_steps\_used + 1$\;
        
        \uIf{$prog.value = \text{TRUE}$}{
            \tcp{Large model successfully unblocks reasoning}
            \If{$\text{is\_final\_answer}(ctx)$}{
                $A \gets \text{extract\_answer}(ctx)$\;
                \Return{$A$}\;
            }
            \tcp{De-escalate back to small model once progress is restored}
            $mode \gets \text{SMALL}$\;
        }
        \ElseIf{$large\_steps\_used \ge B_L$}{
            \tcp{Budget guardrail: avoid unbounded large-model usage}
            $mode \gets \text{SMALL}$\;
        }
    }
}

\tcp{Phase 3: Finalization}
$A \gets \text{extract\_answer}(ctx)$\;
\Return{$A$}\;
\end{algorithm}

The controller in Algorithm~\ref{alg:agentcollab} exposes several desirable properties:

\textbf{Asymmetric Resource Allocation.} The large model is invoked only for the initial warm-up ($K_L$ `Think' steps) and for bounded bursts of escalation (at most $B_L$ `Think' steps per escalation event). In contrast, the small model handles the majority of `Think' steps and tool calls. For typical long-horizon tasks with sparse stagnation regions, this yields a substantial reduction in large-model usage compared to architectures that rely on the large model at every step.

\textbf{Self-Evaluation-Driven Adaptivity.} The Progress Check block provides a concrete, model-generated self-diagnostic signal of whether the trajectory is moving forward. AgentCollab uses this signal to adaptively steer computation: easy tasks and straightforward sub-problems remain on $M_S$, while hard or stuck segments are automatically upgraded to $M_L$ until meaningful progress is restored.

\textbf{Robustness to Stagnation.} Even if the small model frequently fails to make progress, the system can fall back to repeated large-model escalations, trading efficiency for reliability. Conversely, once the large model has re-established a promising direction (indicated by \texttt{prog.value = TRUE}), the controller automatically returns to the small model to keep costs under control.

\textbf{Minimal Computational Overhead.} The self-evaluation progress check is issued immediately after each THINK call and fully reuses the KV cache of the preceding sequence. The generated \texttt{reason}/\texttt{value} fields are typically only 10–20 words, incurring negligible additional decoding latency compared to the substantial savings from avoiding large-model execution on the entire trajectory.

Overall, AgentCollab turns the agent’s own trajectory-level self-evaluation into a \emph{self-governed routing signal} for collaborative inference. By tightly coupling `Think' steps with structured self-evaluation, the system achieves a better balance between end-to-end performance and reasoning quality: simple instances are handled efficiently by the small model, while complex or ill-conditioned cases are elevated to the large model in a targeted, difficulty-aware manner.

\subsection{AgentCompress: Semantic Agent Compression}

This strategy elevates the summarization concept from our case study into a formal, loop-level process. 
We observed that in the workflow of the Information Seeker Agent, the reasoning process typically follows a recurring loop of ``\texttt{think} $\rightarrow$ \texttt{batch\_web\_search} $\rightarrow$ \texttt{url\_crawl} $\rightarrow$ \texttt{document\_qa}'' until the agent determines that the goal has been achieved, at which point it invokes other tools to record results and returns the final conclusion to the Planner Agent. Through profiling and analysis, we identified two major performance bottlenecks:

\begin{enumerate}
    \item \textbf{Parallel overhead in web search and document processing.}
    
    During the batch web search phase, multiple URLs are returned. For each URL, the agent sequentially invokes the {\texttt{url\_crawler}} tool to fetch webpage content and then calls the {\texttt{document\_QA}} module to extract task-relevant information. This introduces a large amount of parallel operations---the combined latency of web-search tools and large-model QA calls significantly slows down and blocks the agent’s workflow.

    \item \textbf{Context bloat during iterative search.}
    As the search process continues, the accumulated web search results can occupy more than 50\% of the total context, with the total sequence length often growing beyond 80K tokens. Such expansion drastically degrades the reasoning efficiency of the main agent.
\end{enumerate}

\begin{figure*}[htbp]
    \centering
    \includegraphics[width=0.85\linewidth]{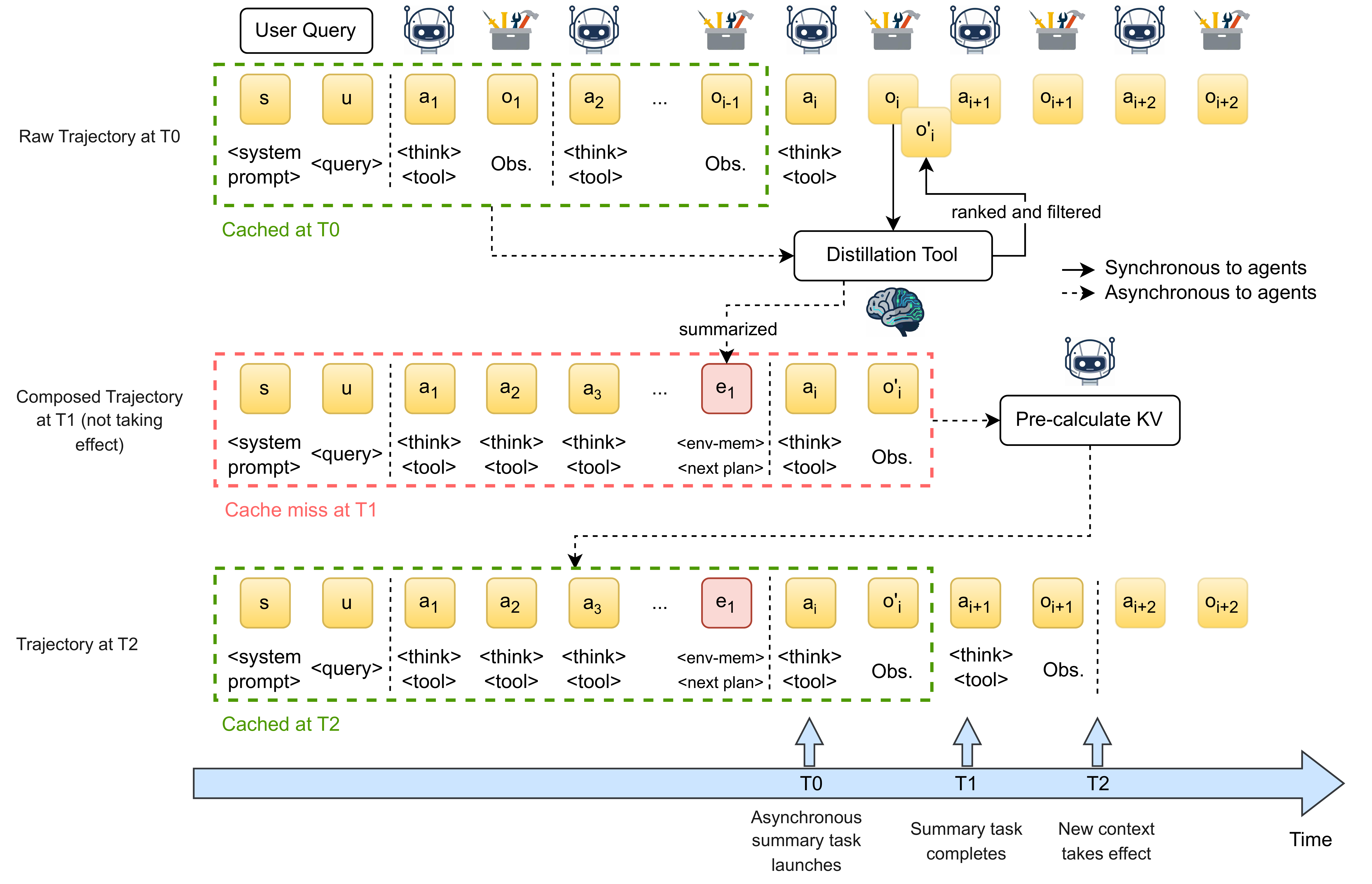}
    \caption{The AgentCompress Framework for Asynchronous Semantic Summarization and Compression.}
    \label{fig:res}
\end{figure*}

To address these issues, we designed and deployed two complementary modules, as shown in Algorithm~\ref{alg:agentcompress}:

\textbf{Search-Enhanced Compression Module.}
After each batch web search, this module invokes a fast lightweight model to filter and rank the retrieved search results. Since the returned results only include snippets and titles, the model can complete ranking and filtering within 5--10 seconds. Irrelevant URLs are pruned and will not proceed to the subsequent url\_crawler or document\_QA stages, effectively reducing tool-level concurrency and improving overall end-to-end performance.

\textbf{Asynchronous Memory Compression and Distillation Module.}
We divide the agent’s \textit{context memory} into two components: \textit{reasoning memory} (from the Think and Act steps) and \textit{environment-interaction memory} (results returned by tools). When the total sequence length exceeds 5K tokens and one search loop is completed, a small model is asynchronously triggered to summarize and distill the current context memory. The goal of this distillation is to transform the existing memory into a structured representation that remains close to the original model’s reasoning process.

After distillation, the reasoning memory is retained, and the distilled context memory is inserted as a new think step immediately following the last compressed step. Since this process is asynchronous, the agent continues reasoning forward during compression. Once the replacement is complete, the next reasoning step adopts the compressed context, allowing compression and reasoning to run in parallel, thereby minimizing end-to-end latency impact.

A notable finding is that retaining reasoning memory is crucial. When only the compressed context memory was kept, the agent experienced cognitive confusion: it lost awareness of its reasoning trajectory and current progress, failed to stop the search loop autonomously, and consequently doubled the number of search iterations. This led to an approximately 1.8$\times$ increase in end-to-end latency, despite the shorter context. In contrast, preserving the reasoning memory maintained the agent’s continuity of thought and situational awareness, effectively preventing reasoning-loop explosion and ensuring stable overall performance.

\begin{algorithm}[H]
\caption{AgentCompress}
\label{alg:agentcompress}
\small
\KwIn{User query $Q$, Agent memory $M$, Search threshold $\theta_{\text{search}}$, Context threshold $\theta_{\text{ctx}}$}
\KwOut{Task result $R$}

\tcp{Initialize working memory}
$M_{\text{reason}} \gets \emptyset$ \tcp{Reasoning memory (think/act steps)}
$M_{\text{env}} \gets \emptyset$ \tcp{Environment interaction memory (tool results)}
$\text{loop\_count} \gets 0$\;

\While{goal not achieved}{
    \tcp{Core reasoning loop}
    $M_{\text{reason}} \gets M_{\text{reason}} \cup \text{Think}(Q, M_{\text{reason}}, M_{\text{env}})$\;
    $\text{urls} \gets \text{batch\_web\_search}(Q, M_{\text{reason}})$\;
    
    \tcp{Search-Enhanced Compression}
    $\text{urls}_{\text{filtered}} \gets \text{LightweightRank}(\text{urls}, Q, \theta_{\text{search}})$\;
    $M_{\text{env}} \gets M_{\text{env}} \cup \text{ParallelProcess}(\text{urls}_{\text{filtered}})$\;
    
    \tcp{Asynchronous Context Compression}
    \If{$|M_{\text{reason}} + M_{\text{env}}| > \theta_{\text{ctx}}$ \textbf{and} loop completed}{
        \tcp{Trigger async compression (non-blocking)}
        $\hat{M}_{\text{env}} \gets \text{AsyncDistill}(M_{\text{env}})$\;
        $M_{\text{env}} \gets \hat{M}_{\text{env}}$ \tcp{Replace after completion}
    }
    
    $\text{loop\_count} \gets \text{loop\_count} + 1$\;
}

\tcp{Generate final result}
$R \gets \text{SynthesizeResult}(M_{\text{reason}}, M_{\text{env}})$\;
\Return{$R$}\;
\end{algorithm}

As shown in Figure~\ref{fig:res}, the summary task runs asynchronously with the agent iteration
while the ranking and filtering of search content blocks the agent until it finishes, so that the following tasks of \texttt{url\_crawl} and \texttt{document\_qa} task will receive more related search results. The KV of the reconstructed context is calculated asynchronously after $T1$ before it takes effect in the agent's iteration, to avoid KV (re-)computation cost due to prefix changes of many originally cached tokens as well as the summarized content.

\subsection{AgentSched: Advanced Agent Scheduler}

Agentic AI inference introduces several distinctive challenges that traditional LLM serving systems are not designed to handle effectively:

\begin{itemize}
    \item \textbf{Multi-turn Conversation Patterns}: Agentic workflows typically involve extended multi-turn interactions where context preservation across turns is critical for maintaining coherent dialogue state and agent memory.
    
    \item \textbf{Extreme Context Length Variance}: Unlike traditional serving with relatively uniform prompt lengths, agentic systems experience orders-of-magnitude variance in context lengths—from short queries ($<$4K tokens) to extended long sequences (32-128K tokens).
    
    \item \textbf{Prefix Recomputation Overhead}: When long-context agents are evicted from the KV cache, the cost of recomputing their extensive prefixes creates significant computational overhead and latency spikes.
\end{itemize}

Therefore, agent inference involves a mixture of long and short sequences. Scheduling must consider both the latency and efficiency of short requests, and the KV cache hit rate and throughput of long, multi-turn conversations. A popular \textbf{First-Come-First-Served (FCFS)} scheduler is ill-suited for this environment, as it can lead to blocking, where a long, computationally heavy request at the front of the queue stalls numerous short, latency-sensitive requests behind it. Furthermore, indiscriminate scheduling can thrash the KV cache, evicting valuable context from long-running agents and forcing costly recomputation of their prefixes in subsequent turns.

\begin{figure}[!htp]
    \centering
    \includegraphics[width=0.85\linewidth]{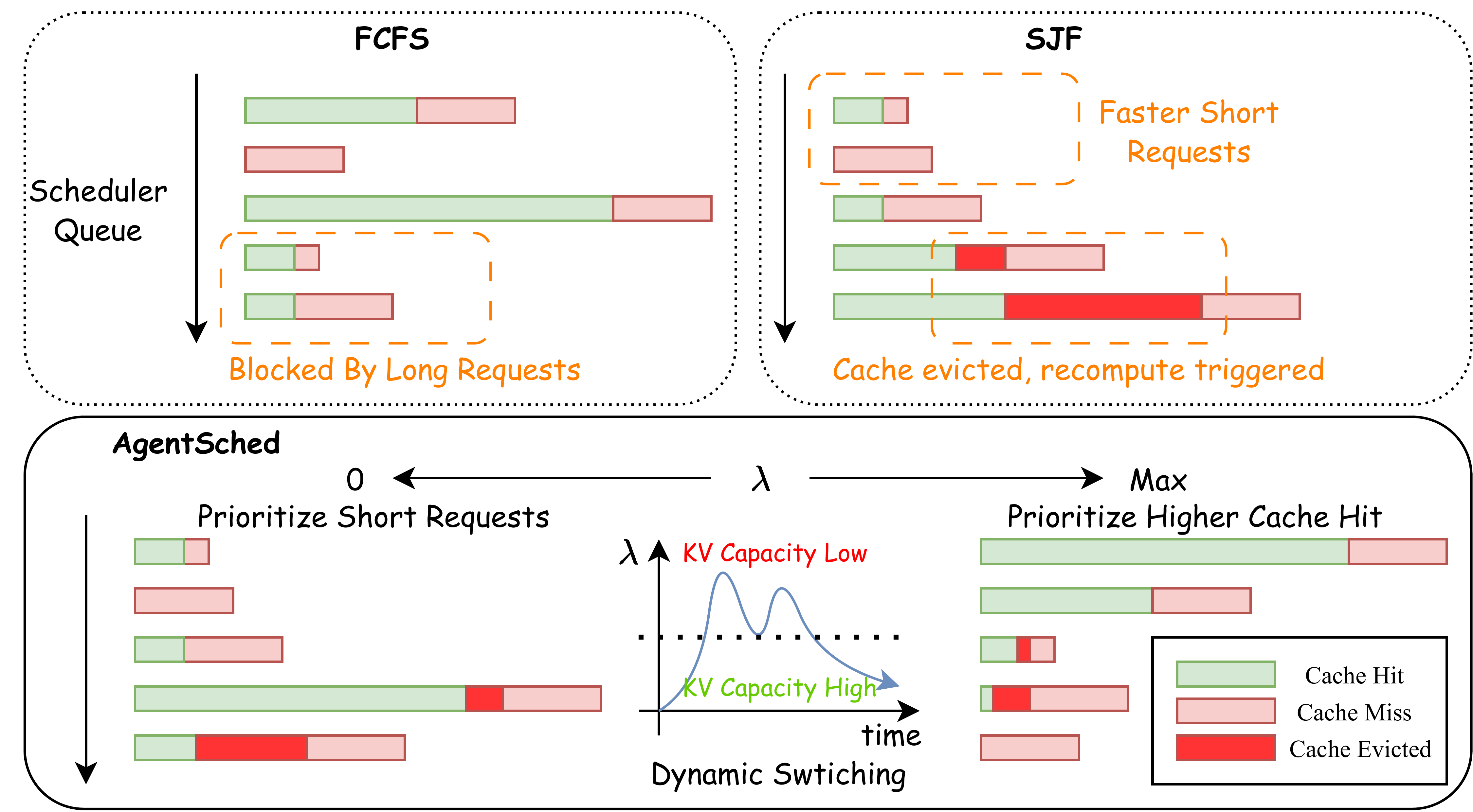}
    \caption{Overview of AgentSched. AgentSched uses a controller signal $\lambda$ to dynamically switch between a Shortest-Job-First (SJF) mode and a KV-aware mode that prioritizes higher cache-hit sequences, achieving a better trade-off between latency and cache reuse in agent scheduling scenarios with a mix of very long and short sequences.}
    \label{fig:agentsched_overview}
\end{figure}

A more proactive strategy is \textbf{Shortest-Job-First (SJF)}, a classic scheduling algorithm that prioritizes requests with the smallest estimated processing time. In the context of LLM inference, ``job length'' can be approximated by the number of tokens in the prompt (for prefill).
SJF is effective at reducing average job latency; by running short jobs first, it clears the queue rapidly, reducing the average TTFT.
Shorter requests, which are usually tool agents (such as \texttt{document\_qa}), can be scheduled earlier by SJF to keep the workflow running and avoid blocking their parent executor agents.

However, a pure SJF policy can be detrimental in agentic scenarios. It may consistently deprioritize long-running agents, leading to starvation. More critically, it ignores the KV cache state. Scheduling a short, new request with no KV cache hits over a long, cached agent can evict the agent's context, resulting in a high ``KV cache miss'' penalty when that agent is next scheduled, undermining overall system efficiency. 

\begin{algorithm}[H]
\caption{Agent Scheduler}
\label{alg:agentscheduler}
\small
\KwIn{Free KV blocks $N$, request queue $\mathcal{Q}$}
\KwOut{Selected request $picked$}

\tcp{Step 0: Initialize parameter}
Shadow price $\lambda$, integral state $I$\;
Parameters $(\lambda_{\max}, k, \varepsilon, k_P, k_I, \beta)$\;
Block size $tpb$\;

\tcp{Step 1: Preprocess}
Periodically refresh per-request blocks hit and new blocks needed\;
$hit_{i}$, $need_{i}$\;

\tcp{Step 2: Update Shadow Price $\lambda$}
\tcp{Compute effective capacity}
$H \gets \sum_{i \in \mathcal{Q}} hit_i$\;
$U \gets \max(0,\, N - H)$\;
$D \gets \sum_{i \in \mathcal{Q}} need_i$\;

$gap^\ast \gets D - U$\;
$z \gets \frac{D}{U + \varepsilon}$\;

\tcp{Soft mapping of pressure to $\lambda$}
$\lambda \gets \lambda_{\max} \cdot \sigma\!(k(z - 1))$\;

\tcp{Step 3: Compute Score for Each Request}
\ForEach{$i \in \mathcal{Q}$}{
    ${\mathit{prompt\_tok}} \gets \text{len(prompt of }i)$\;
    ${\mathit{need\_tok}} \gets {\mathit{prompt\_tok}} / {\mathit{tpb}}$\;

    ${\mathit{sjf\_mix}} \gets \text{clamp}(1 - \lambda / \lambda_{\max},\, 0,\, 1)$\;
    ${\mathit{need\_eff}}_i \gets {\mathit{sjf\_mix}} \cdot {\mathit{need\_tok}} + (1 - {\mathit{sjf\_mix}}) \cdot need_i$\;

    \tcp{Score = hit reward - KV demand penalty + wait reward}
    $score_i \gets a \cdot hit_i - (b + \lambda)\cdot {\mathit{need\_eff}}_i + c \cdot wait_i$\;
}
$candidates \gets \text{sort\_by\_descending\_score}(\mathcal{Q})$\;

\tcp{Step 4: Find Feasible Primary Candidate}
$picked \gets \text{first\_feasible}(candidates)$\;
\Return{$picked$}\;
\end{algorithm}

\textbf{Hybrid Scheduling.}
To capture the benefits of SJF while protecting system efficiency, we employ a hybrid, adaptive approach: \textbf{AgentSched}. Its core innovation is a shadow-price (\(\lambda\)) mechanism that dynamically measures the pressure on the KV cache capacity and adjusts the scheduling policy.

The scheduler continuously monitors two key quantities:
\textbf{Effective Usable Capacity (\(U\))}: The number of free KV cache blocks, adjusted for blocks that are ``reserved'' because they can be reused (hit) by requests in the queue. \textbf{Total New Demand (\(D\))}: The sum of new KV cache blocks required by all requests in the queue.

The relationship between \(D\) and \(U\) defines the system's state. When \(D \leq U\) (loose capacity), the system can comfortably accommodate all new demand. When \(D > U\) (tight capacity), demand exceeds capacity, and scheduling decisions critically impact which requests' cached contexts get evicted.

The shadow-price \(\lambda\) translates this binary state into a continuous control signal. It is computed using a smooth function (e.g., a sigmoid) of the ratio \(D/U\):

\textbf{Loose KV Cache (\(\lambda \rightarrow 0\))}: The scheduler behaves like a standard SJF policy, prioritizing requests with short prompts to minimize average latency.

\textbf{Tight KV Cache (\(\lambda \rightarrow high\))}: The scheduler smoothly transitions to a KV-aware mode. The score function used to sort requests increasingly penalizes the allocation of \emph{new} cache blocks (\textit{need}) and rewards requests with high cache \textit{hit} counts.

This results in a priority score for each request \(i\) of the form: $\text{score}_i = a \times \text{hit}_i - (b + \lambda) \times \text{need\_eff}_i + c \times \text{wait}_i$. 
Here, \(\text{need\_eff}_i\) is a blend of token-based and block-based need, controlled by \(\lambda\), ensuring a smooth transition between SJF and KV-aware behavior.
$c \times \text{wait}_i$ is a term that prioritizes sequences that have waited for long. 


In summary, advanced scheduling for Agentic AI moves beyond simple policies to adaptive, hybrid algorithms. By dynamically balancing the competing goals of low latency and high cache efficiency, systems can achieve robust performance across the diverse workload mix characteristic of agentic inference.

\subsection{AgentSAM: Context Memory Enhanced SAM Speculative Decoding} 

\subsubsection{Methodology}
In agent inference, requests derived from the same user query—as well as those from different users with similar intents—exhibit strong correlations among the planner agent, executor agents, and tool agents. In deep research settings, retrieved search content is frequently revisited during the model’s reasoning process, making such environments well-suited for lightweight n-gram–based speculative decoding algorithms. We employ suffix automaton (SAM)~\cite{hu2024samdecodingspeculativedecoding} as a lightweight, hot-swappable speculative mechanism designed to provide lossless acceleration for small-batch and long-tail tasks.

\subsubsection{Algorithm design} 
SAM speculative decoding proposes indexing and looking up all substrings of the context in a suffix automaton (SAM). The request prompt is first used to build the SAM. During decoding, after verifying the tokens with the LLM, given the current verified suffix, the decoder follows SAM transitions to retrieve a matching continuation with a specified length, which is emitted as a batch of speculative tokens. Verified tokens are then inserted into the SAM to keep the index consistent with the growing context, while any mismatches trigger fallback to standard decoding. This design provides an efficient, n-gram-style mechanism for low-overhead speculation and validation.

To further enhance the effectiveness of SAM in agent inference scenarios, we propose \textbf{AgentSAM}, which leverages both user/session-level context memory and cross-session context memory to improve speculative hit rates.
AgentSAM maintains a centralized repository of contextual memory—comprising prompts and corresponding LLM responses from past agent interactions across all users and sessions. When processing a new request, the system constructs a composite suffix automaton by integrating two sources of contextual information:
(1) the context associated with the current user session (i.e., prior turns with planner agents and other executor agents in the same session), and
(2) the top-$K$ most semantically similar historical contexts retrieved via dense and/or sparse retrieval methods based on similarity scores between the incoming query and stored queries of other sessions.
Finally, the SAM structures from these selected contexts are merged to form a unified SAM, to which the current prompt is appended. The similarity scores determine the weights of each SAM during the merge.
This enriched automaton then guides the drafting phase of speculative decoding. By initializing SAM with contextually relevant tokens drawn from prior agent interactions that exhibit strong semantic affinity to the current query, AgentSAM achieves a higher speculative hit rate, thereby improving the overall efficiency and end-to-end performance of speculative decoding.

\subsubsection{System design}
In order to optimize the inference performance, we developed a system design for speculative decoding in agentic inference.

\textbf{Asynchronous SAM construction.} Suffix automaton is constructed incrementally token by token; consequently, the computation cost of building a SAM scales linearly with prompt length. For long prompts in agentic tasks (up to 128k tokens), the construction latency can reach on the order of one second. The construction of the SAM before decoding stage will substantially increase TTFT, compromising the time cost saved from speculative decoding. To address this, we offload the SAM construction to a child thread. Upon receiving a new request, the runtime attempts to retrieve the corresponding SAM; if the SAM is not yet available, the prompt or any generated tokens are put into a waiting queue, and speculative drafting is temporarily disabled for that request. The child thread continuously examines the waiting queue and constructs SAM on demand according to the queue entries. By decoupling SAM construction from the main inference thread, this design prevents SAM construction from blocking the low-latency decoding path and preserves the responsiveness required for effective speculative decoding.

\textbf{Adaptive switch for speculative decoding.} Speculative decoding leverages the excessive compute available by performing drafting and then validating additional speculative tokens; the realized performance improvement therefore depends critically on (1) the validation overhead for each speculative token being small relative to the savings from avoiding full-model decoding, and (2) a sufficiently high speculative hit rate. Because the likelihood of repeated context patterns—and hence the effectiveness of lightweight, n-gram–based drafting model such as SAM—varies substantially across application scenarios, the achievable hit rate for such methods is highly workload-dependent. Consequently, for any given deployment there exist practical thresholds on context length and batch size beyond which speculative decoding ceases to provide net benefit. To address this, we implement an adaptive speculative decoding switch: the runtime monitors context length and batch size and disables speculative decoding (falling back to standard, non-speculative inference) whenever those parameters exceed empirically determined, system-specific thresholds, thereby avoiding wasted validation effort in regimes where speculation is unlikely to pay off.

\begin{algorithm}[H]
\caption{AgentSAM}
\label{alg:agentsam}
\small
\KwIn{Incoming request $q$, session context $C_{\text{session}}$, global memory $M_{\text{global}}$}
\KwOut{Generated response $y$}

\tcp{1. Retrieve and prepare context}
$S_{\text{session}} \gets \text{BuildSAM}(C_{\text{session}})$\; 
$S_{\text{cross}} \gets \text{RetrieveTopK}(q, M_{\text{global}})$\;  
$S_{\text{merged}} \gets \text{MergeSAM}(S_{\text{session}}, S_{\text{cross}}, \text{weights})$\;

\tcp{2. Asynchronous SAM construction}
Launch background thread to incrementally extend $S_{\text{merged}}$ with streamed tokens\;

\tcp{3. Speculative decoding loop}
\While{not done}{
    $x_t \gets$ current verified suffix\;
    $\hat{x}_{t:t+k} \gets \text{FollowTransitions}(S_{\text{merged}}, x_t)$\;
    Emit speculative tokens $\hat{x}_{t:t+k}$\;
    Validate with main LLM; commit matched prefix\;
    \If{mismatch detected}{
        Fallback to standard decoding\;
    }
}

\tcp{4. Adaptive controller}
\If{context length $>$ threshold $\lor$ batch size $>$ threshold}{
    Disable speculation; use normal decoding\;
}
\Else{
    Continue SAM-based speculation\;
}

\Return{$y$}\;
\end{algorithm}


\section{Experimental Evaluation}

\subsection{Experimental Setup}
\begin{itemize}
    \item \textbf{Tasks:} Experiments were conducted on the \textbf{BrowseComp-zh (Chinese version)}~\cite{zhou2025browsecomp} benchmark, which involves multi-step tasks requiring web searches.
    \item \textbf{Models:} The primary model used was openPangu-DeepDiverV2-7B~\cite{openpangu_deepdiver_v2}, with openPangu-7B~\cite{openpangu-embedded-7b} used for context compression in AgentCompress and openPangu-DeepDiverV2-38B~\cite{openpangu_deepdiver_v2} used for the large model in AgentCollab.
    \item \textbf{Infrastructure:} vLLM-Ascend v0.9.1~\cite{vllm-ascend} on Ascend 910B3 NPUs.
\end{itemize}

\subsection{Evaluation on AgentCollab}

We first evaluate AgentCollab on the BrowseComp-zh benchmark, comparing it with a large-model-only agent and a small-model-only agent. 
Table~\ref{tab:agentcollab_overall} reports task accuracy and normalized end-to-end performance (larger is better, large-model-only set to $1.0\times$).

\begin{table}[htbp]
\caption{Overall accuracy and normalized end-to-end performance on \textsc{BrowseComp-zh}.}
\label{tab:agentcollab_overall}
\centering
\small
\begin{tabular}{lcc}
\toprule
\textbf{Method} & \textbf{Acc. (\%)} & \textbf{Speedup} \\
\midrule
Large-model only & 34.6  & $1.00\times$ \\
Small-model only & 18.3  & $1.54\times$ \\
- w/ large-model as Planner & 24.6  & $1.39\times$ \\
- w/ large-model as InfoSeeker & 27.3  & $1.24\times$ \\
AgentCollab      & 33.8  & $1.32\times$ \\
\bottomrule
\end{tabular}
\end{table}

From an accuracy perspective, the small-model-only agent suffers a substantial degradation: its accuracy drops to $18.3\%$, losing nearly half of the large model’s performance.
In contrast, AgentCollab attains $33.8\%$ accuracy, recovering most of the large-model-only performance while using the large model only when its self-evaluation signal indicates difficulty.
In other words, relative to the small model, AgentCollab delivers almost a $85\%$ gain in accuracy (33.8\% vs.\ 18.3\%) without reverting to a fully large-model pipeline.

On the efficiency side, the small-model-only agent achieves a $1.54\times$ end-to-end speedup over the large-model-only baseline, but at the expense of severe quality loss.
AgentCollab provides a $1.32\times$ end-to-end speedup while maintaining near-identical accuracy to the large model.
This demonstrates that self-evaluation-driven escalation can move the operating point of the agent closer to the Pareto frontier of accuracy–efficiency trade-offs.

\begin{table*}[htbp]
\centering
\caption{Comparison of key THINK steps across three reasoning modes: small-model-only, AgentCollab, and large-model-only.}
\label{tab:agentcollab_case_study}
\small
\begin{tabular}{p{0.3\linewidth} p{0.3\linewidth} p{0.3\linewidth}}
\toprule
\textbf{Small Model} &
\textbf{AgentCollab} &
\textbf{Large Model} \\
\midrule

\textcolor{blue}{
Let me analyze this task: The task requires searching for a historically significant document jointly signed by multiple national representatives, with a focus on the English copy stored in a Chinese museum. I should begin with a web search.
}
&
\textcolor{blue}{
Let me analyze this task: I need to search for a multi-nation jointly signed historical document, stored in a Chinese museum, in English. Possible candidates include the UN Charter, Treaty of Versailles, etc. I need both Chinese and English keywords.
}
&
\textcolor{blue}{
Let me analyze this task: Find an English document jointly signed by multiple national representatives with major historical significance, and stored in a Chinese museum. I should focus on international treaties, UN documents, and museum collections in China.
}
\\

\multicolumn{3}{p{0.95\linewidth}}{
\textbf{Analysis:}
All three systems begin with high-quality reasoning from the large model. \textit{The Small Model mode also starts with large model planning to compare them in a more fair configuration.}
}
\\
\midrule
......
&
......
&
......
\\
\midrule

\textcolor{red}{
Now I have obtained detailed information on the Cairo Declaration… Let me search for other similar important documents.
}
&
\textcolor{red}{
I still have not found a multi-nation signed English document stored in a Chinese museum. Let me try searching for the “UN Conference Memorial Album” from the Museum of Overseas Chinese History.
}
\textcolor{orange}{
\textbf{[Progress check: no significant progress made. Escalate.]}
}
&
\textcolor{blue}{
I found important information: UN Charter drafting documents (Chongqing Library), “UN Conference Memorial Album” (Museum of Overseas Chinese History), and other multi-nation historical materials.
}
\\

\multicolumn{3}{p{0.95\linewidth}}{
\textbf{Analysis:}
The left system enters a local loop with limited search breadth; the AgentCollab system encounters stagnation and triggers model escalation; the right system continues high-quality knowledge expansion.
}
\\
\midrule
......
&
......
&
......
\\
\midrule
\textcolor{red}{
From the second round of search results, I did not find other documents that clearly meet the requirements… Let me search the UN Charter or other international treaties.
}
&
\textcolor{blue}{
Great! I found the “UN Conference Memorial Album”: stored in the Museum of Overseas Chinese History, involving 50 nations, containing bilingual UN Charter texts, fully meeting the task requirements.
}
\textcolor{orange}{
\textbf{[Restore to small model.]}
}
&
\textcolor{blue}{
Summary of findings: English UN Charter, UN Conference Memorial Album, “Records of Liberated Areas of China,” and other relevant documents.
}
\\

\multicolumn{3}{p{0.95\linewidth}}{
\textbf{Analysis:}
This is where collaborative reasoning shows its clear advantage: the left system fails to find the key document; the middle system, after escalation, finds the exact same crucial evidence as the large-model-only system.
}
\\
\midrule
......
&
......
&
......
\\
\midrule
\textcolor{red}{
Now I have enough information to complete the task: Cairo Declaration, UN Charter (museum collection unclear).
}
&
\textcolor{red}{
I found three classes of documents: the UN Conference Memorial Album; UN documents stored in the National Library of China; and international convention files stored through UN depository library systems (including English versions).
}
&
\textcolor{blue}{
Let me reorganize all findings: English UN Charter, UN Conference Memorial Album, Records of Liberated Areas of China, and UN Charter drafting documents — ready to finish the task.
}
\\

\multicolumn{3}{p{0.95\linewidth}}{
\textbf{Analysis:}
The right system maintains maximal coverage at high cost; the middle system reaches nearly the same breadth and accuracy through collaboration; the left system misses the core target and yields the lowest final quality.
}
\\

\bottomrule
\end{tabular}
\end{table*}

A trajectory-level case study in Table~\ref{tab:agentcollab_case_study} further illustrates how these aggregate numbers arise in practice.
In the small-model-only setting, the agent easily falls into local search loops (e.g., repeatedly exploring variants of the \emph{Cairo Declaration}) and fails to locate the key document.
Under AgentCollab, the self-evaluation (\emph{Progress Check}) correctly flags such segments as non-progressive, triggering escalation to the large model, which broadens the search space and quickly recovers the crucial evidence (e.g., the “UN Conference Memorial Album”).
After progress is restored, control returns to the small model to complete the remaining reasoning and summarization.
The large-model-only agent follows a similarly strong trajectory but pays a higher computational cost throughout.
Taken together, these results show that AgentCollab can match the reasoning quality of a large-model agent on web-heavy tasks, while delivering substantially better end-to-end efficiency through self-evaluation-driven collaboration.

\subsection{Evaluation on AgentCompress}

We evaluate Search Compression and Context Compression of AgentCompress on BrowseComp-zh.
As shown in Tables~\ref{tab:accuracy_results_compression} and~\ref{tab:latency_results_compression}, AgentCompress achieves substantial token reduction—compressing context from over 40K tokens to around 20K—while preserving task accuracy and significantly lowering end-to-end latency.

\begin{table}[htbp]
\centering
\caption{Accuracy (\%) of AgentCompress. AgentCompress preserves the end-to-end accuracy.}
\label{tab:accuracy_results_compression}
\begin{tabular}{lc}
\toprule
\textbf{Method} & \textbf{Accuracy (\%)} \\
\midrule
Baseline (Google Search) & 18.30 \\
+ AgentCompress (Search Compression)  & 21.80 \quad ($\uparrow$3.50\%) \\
+ AgentCompress (Context Compression)  & 19.03 \quad ($\uparrow$0.73\%) \\
\bottomrule
\end{tabular}
\end{table}

\begin{table}[htbp]
\centering
\caption{Comparison of tool call average count and latency in one session with Search Compression. (count/average latency(s))}
\label{tab:tool_call_search_compression}
\begin{tabular}{lcc}
\toprule
\textbf{Tool Call} & \textbf{Baseline} & \textbf{Search Compression} \\
\midrule
Batch Web Search & 19.59 / ~3.27 & 20.142 / 12.77 \\
Url Crawler & ~6.86 / 10.37 & ~6.44 / ~9.20 \\
Document QA & ~6.47 / 17.55 & ~5.89 / 10.88 \\
\bottomrule
\end{tabular}
\end{table}

\textbf{Search Compression.}
Experimental statistics reveal that token counts from search content constitute 50\% or more of the final agent trajectory context. Re-ranking and filtering search content can eliminate irrelevant materials (e.g., advertisements) in user query scenarios, accounting for approximately 20\% of search content tokens. This reduces context length despite introducing around 10 seconds of processing overhead per search function call. The refined, higher-quality search results minimize content processing in downstream \texttt{url\_crawl} and \texttt{document\_qa} stages while maximizing avoidance of irrelevant content extraction. 
Table~\ref{tab:accuracy_results_compression} shows that Search Compression leads to an accuracy increase of 3.50\% when using the Google Serper API, suggesting that the filtering process could potentially improve the end-to-end accuracy of the system.
Experiments in Table~\ref{tab:tool_call_search_compression} demonstrate average latency reductions of 11.2\% and 38.0\% for \texttt{url\_crawler} and \texttt{document\_qa} respectively compared to baseline, with invocation counts decreasing by 6.12\% and 8.9\%. The batch-web-search tool call latency includes LLM call time, which is around 10s. Moreover, as this compression method preserves historical context integrity, it avoids cache miss issues.
We conclude that the Search Compression module enhances overall system performance in terms of latency, without compromising accuracy.

\textbf{Context Compression.}
Table~\ref{tab:accuracy_results_compression} shows that Context Compression yields a slight improvement (+0.73\%) with Google Search due to noise reduction in retrieved context. This confirms that well-designed summarization can enhance system performance without sacrificing fidelity.

\begin{table}[htbp]
\centering
\caption{Latency reduction and turn count with AgentCompress}
\label{tab:latency_results_compression}
\adjustbox{max width=\linewidth}{
\begin{tabular}{lcc}
\toprule
\textbf{Model Configuration} & \textbf{Latency Reduction} & \textbf{\# Turns} \\

\midrule

Baseline (TP=1, $N_{\text{parallel}}$=1) & 0\% & 1.0$\times$ \\
w/ AgentCompress (Search Compression) & -9.8\% & 1.02$\times$ \\
w/ AgentCompress (Context Compression) & –6.1\% & 1.04$\times$ \\

\midrule

Baseline (TP=1, $N_{\text{parallel}}$=4) & 0\% & 1.0$\times$ \\
w/ AgentCompress (Search Compression) & -22.1\% & 0.97$\times$ \\
w/ AgentCompress (Context Compression) & –23.4\% & 1.03$\times$ \\

\midrule

Baseline (TP=1, $N_{\text{parallel}}$=8) & 0\% & 1.0$\times$ \\
w/ AgentCompress (Search Compression) & -13.2\% & 1.01$\times$ \\
w/ AgentCompress (Context Compression) & –42.1\% & 1.08$\times$ \\

\midrule

Baseline (TP=2, $N_{\text{parallel}}$=1) & 0\% & 1.0$\times$ \\
w/ AgentCompress (Search Compression) & -12.6\% & 1.03$\times$ \\
w/ AgentCompress (Context Compression) & –6.4\% & 1.03$\times$ \\

\midrule

Baseline (TP=2, $N_{\text{parallel}}$=4) & 0\% & 1.0$\times$ \\
w/ AgentCompress (Search Compression) & -12.1\% & 0.98$\times$ \\
w/ AgentCompress (Context Compression) & –7.0\% & 1.07$\times$ \\

\midrule

Baseline (TP=2, $N_{\text{parallel}}$=8) & 0\% & 1.0$\times$ \\
w/ AgentCompress (Search Compression) & -8.5\% & 0.96$\times$ \\
w/ AgentCompress (Context Compression) & –9.0\% & 1.06$\times$ \\

\bottomrule
\end{tabular}}
\end{table}

As demonstrated in Table~\ref{tab:latency_results_compression}, AgentCompress consistently reduces end-to-end execution time across diverse tensor-parallelism and batching configurations. The most dramatic speedup (42.1\%) occurs under high-batch, single-tensor-parallelism settings (TP=1, $N_\text{parallel}$=8), where memory pressure is highest—highlighting AgentCompress’s role in alleviating HBM bottlenecks. Even in low-latency regimes (e.g., TP=2, $N_\text{parallel}$=1), we observe 6–12\% improvements.

\begin{table}[htbp]
\centering
\caption{Ablation study on context components after context compression.}
\label{tab:ablation_context}
\begin{tabular}{lcc}
\toprule
\textbf{Composition Variations} & \textbf{End-to-End Latency} & \textbf{Number of Turns} \\
\midrule
Baseline & 0\% & 1.0$\times$ \\
Proposed Composition & -6.4\% & 1.07$\times$ \\
- reasoning traces & +31.98\% & 1.12$\times$ \\
- summary & +21.32\% & 1.09$\times$ \\
- recent tool responses  & +29.50\% & 1.26$\times$ \\
\bottomrule
\end{tabular}
\end{table}

Context Compression is triggered at the conclusion of each search round, identified by a ``search function call'' from the agent. To maintain coherence, summaries retain both recent tool responses and the agent’s reasoning traces; ablation studies in Table~\ref{tab:ablation_context} confirm that omitting either increases dialogue turns by up to 26\%. 
The proposed composition achieves a 6.4\% reduction in end-to-end latency while maintaining a comparable number of conversational turns, indicating improved efficiency without compromising interaction quality. In contrast, removing key contextual elements such as reasoning traces, summaries, or recent tool responses leads to substantial latency increases (ranging from +21.32\% to +31.98\%) and higher turn counts, suggesting that these components are essential for preserving reasoning continuity and reducing redundant dialogue iterations. These results highlight that effective compression is not solely about minimizing context length but about selectively retaining semantically rich information that supports coherent reasoning and efficient task completion.

Summaries are generated asynchronously ($\approx$20 s latency) using a concise prompt template aligned with the agent’s native reasoning format, ensuring minimal interference with ongoing inference. 
The updated context only takes effect at the end of a subsequent round after computation completes. 
From initiating the summary request to generating the summary and precomputing the new KV cache, the agent typically progresses through 2 or 3 rounds. Subsequent LLM inference latency benefits from the shorter context with almost no cache miss. 
On the other hand, Table~\ref{tab:latency_results_compression} shows compression strategies slightly increase agent trajectory turns, reducing potential performance gains. This turn increase may stem from information loss in summaries or distribution shifts between training and inference data. 
However, a noticeable decrease in end-to-end latency is still observed, primarily resulting from the reduction in context length.

In summary, AgentCompress delivers a favorable trade-off: it reduces token consumption by over 50\%, cuts end-to-end latency by 6–42\% depending on system load, and maintains or slightly improves task accuracy—all while operating transparently within the agent’s cognitive loop.

\subsection{Evaluation on AgentSched}

We evaluate AgentSched against standard scheduling baselines on three key performance metrics: average LLM call latency, end-to-end latency, and average KV cache hit rate. The test configuration is: a 7B model deployed with Tensor Parallel (TP) = 2. Max sequence length is 128K, and the number of parallel agent sessions is 4. Under this setting, the KV Cache will become insufficient when agent sessions start processing information seeking tasks that requires 10-30 iterations.

\begin{table*}[htbp]
\centering
\caption{Performance comparison of scheduling strategies}
\label{tab:performance_comparison}
\begin{tabular}{lccc}
\toprule
\textbf{Scheduling Method} & \textbf{Avg. LLM Call Latency} & \textbf{End-to-End Latency} & \textbf{Avg. KV Cache Hit Rate} \\
\midrule
FCFS (Baseline) & 1.0$\times$ & 1.0$\times$ & 63\% \\
SJF & 0.948$\times$ & 0.952$\times$ & 58\% \\
AgentSched  & 0.902$\times$ & 0.904$\times$ & 72\% \\
\bottomrule
\end{tabular}%
\end{table*}

\begin{figure}[!ht]
    \centering
    \includegraphics[width=0.7\linewidth]{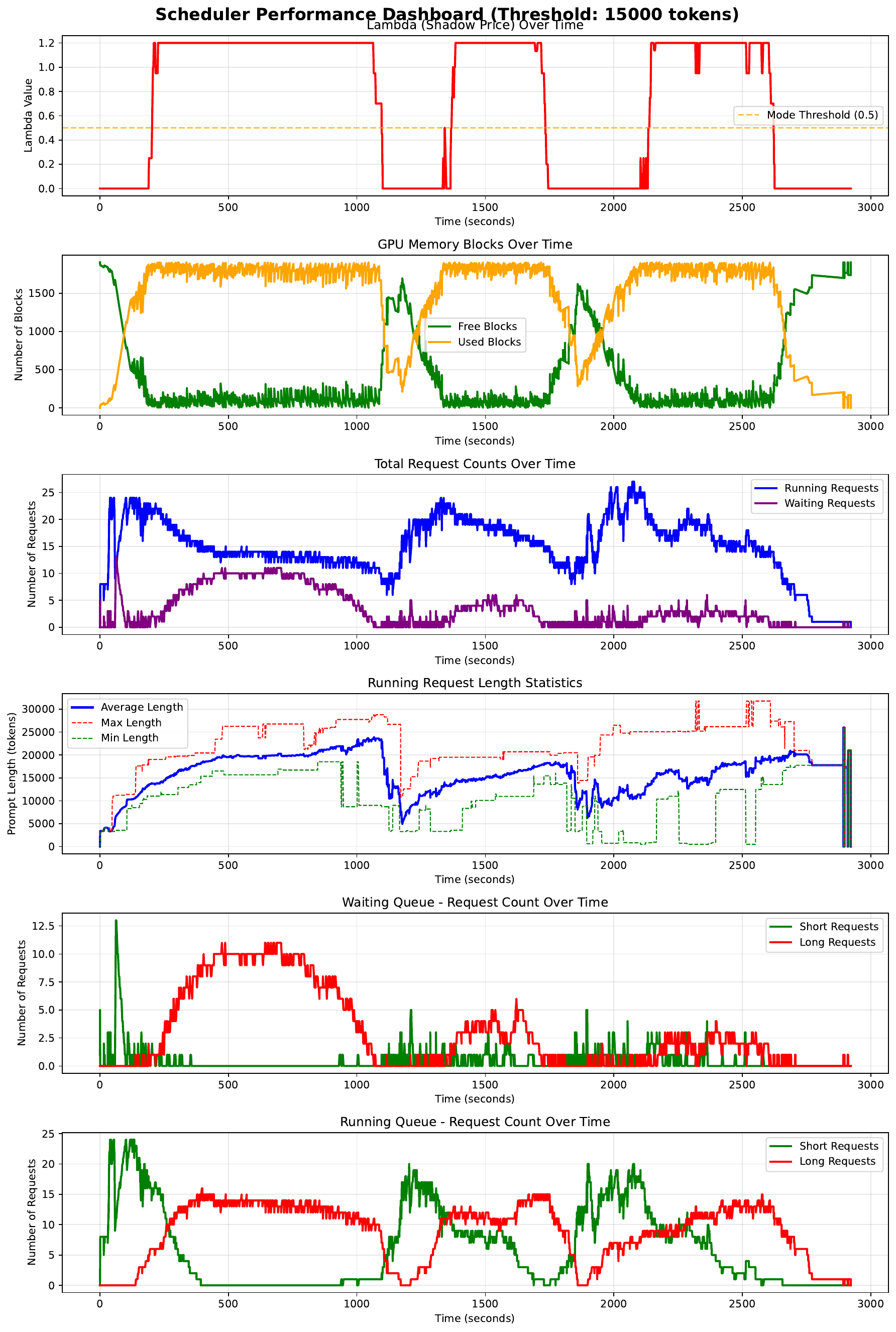}
    \caption{System metrics during the scheduling process. Under different workloads, the controlling factor $\lambda$ smoothly transists between two modes, leading to dynamic scheduling mode that balances request throughput and KV cache hit rate.}
    \label{fig:agent_scheduler_demo}
\end{figure}

\textbf{Discussion.}
For better visualization, all incoming requests are categorized into either \emph{long requests} or \emph{short requests}, based on whether the prompt length exceeds 10k tokens. The results in Figure~\ref{fig:agent_scheduler_demo} demonstrate that:
\begin{itemize}
    \item When $\lambda$ is small ($\lambda < 0.5$), a greater number of short requests are admitted into the \texttt{running queue} for execution. As a result, the total number of requests in the \texttt{running queue} increases.
    \item When $\lambda$ is large ($\lambda > 0.5$), more long requests are scheduled into the \texttt{running queue}, which consequently reduces the overall number of requests in the queue due to their longer execution times.
\end{itemize}
This classification and scheduling mechanism ensures that system throughput and responsiveness are optimized according to current resource availability.

The results in Table~\ref{tab:performance_comparison} demonstrate that AgentSched achieves consistent improvements across both latency and cache efficiency dimensions. Specifically, AgentSched reduces average LLM call latency to 0.902$\times$ and end-to-end latency to 0.904$\times$ of the FCFS baseline—corresponding to approximately 9.8\% and 9.6\% reductions, respectively—while simultaneously increasing the average KV cache hit rate from 63\% to 72\%, a relative improvement of 14.3\%.

In contrast, SJF yields only marginal latency benefits (4.8–5.2\% reduction) but degrades cache performance, lowering the hit rate to 58\%. This reflects SJF’s tendency to prioritize short requests at the expense of context continuity, leading to more frequent KV cache evictions for longer or multi-turn agent sessions.

AgentSched avoids this trade-off by integrating context-aware scheduling decisions that preserve active agent states while still accelerating interactive turns. The concurrent reduction in latency and increase in cache hit rate underscore the effectiveness of its adaptive mechanism in aligning scheduling priorities with the memory-access patterns of agentic workloads. These results confirm that AgentSched not only outperforms traditional policies in isolation but also reconciles objectives that are typically in tension—namely, low latency and high cache reuse.

\subsection{Evaluation on AgentSAM} 

At each decoding step, AgentSAM generates a proposal set of $N_{\text{propose}} = 4$ tokens: one token produced by standard autoregressive decoding and three additional tokens predicted speculatively using AgentSAM.
We construct cross-session trajectories by leveraging generated sequences from all other questions in BrowseComp-zh, excluding the current one.

To quantify acceleration effectiveness, we consider three metrics:
\begin{itemize}
    \item \textbf{End-to-End Latency (E2E)}: the end-to-end time cost for completing each session.
    \item \textbf{Overall Token Efficiency (OTE)}: the ratio of total generated tokens to the number of actual model forward passes during decoding. This reflects the end-to-end computational efficiency across the entire generation session.
    \item \textbf{Speculative Hit Rate (SHR)}: the proportion of accepted speculative tokens among the $N_{\text{propose}} - 1$ proposed tokens (excluding the base token), conditioned on AgentSAM issuing a non-empty speculative proposal. This measures the accuracy and utility of the speculative predictions when they are attempted.
\end{itemize}

\begin{table}[htbp]
\centering
\caption{Ablation study of end-to-end latency reduction after applying AgentSAM, TP=2.}
\label{tab:e2e_latency_sam}
\adjustbox{max width=\linewidth}{
\begin{tabular}{lcc}
\toprule
\textbf{Model Configuration} & \textbf{End-to-End Latency} & \textbf{LLM Latency} \\

\midrule

Baseline (No speculative decoding) & 0 \% & 0 \% \\
SAM speculative decoding & -16.3 \% & -20.7 \%  \\
AgentSAM speculative decoding & -21.2 \% & -26.0 \%  \\

\midrule
\end{tabular}}
\end{table}


OTE captures the holistic speedup achieved by the system, while SHR isolates the quality of AgentSAM’s speculative proposals independent of scheduling or fallback behavior.

\textbf{End-to-End Latency Discussion.}
Our experiments show that speculative decoding integrated into the agent’s LLM inference pipeline yields substantial end-to-end performance benefits. Under a configuration of tensor parallelism = 2 and batch size = 4, employing SAM-based speculative decoding built from the request’s contextual tokens reduces the LLM latency by approximately 20.7\%. When accounting for non-LLM components of the agent workflow—such as tool calls, retrieval steps, and environment interactions—the overall end-to-end latency reduces by about 16.3\%. Furthermore, by leveraging AgentSAM, the speculative hit rate increases noticeably. This improvement translates to an LLM latency reduction of up to 26\%, ultimately yielding an end-to-end latency reduction of roughly 21.2\%. These results highlight that speculative decoding not only accelerates the core inference loop but also provides tangible gains at the full agent-system level, where cumulative latencies amplify the value of each millisecond saved during generation.

\begin{figure}
    \centering
    \includegraphics[width=0.6\linewidth]{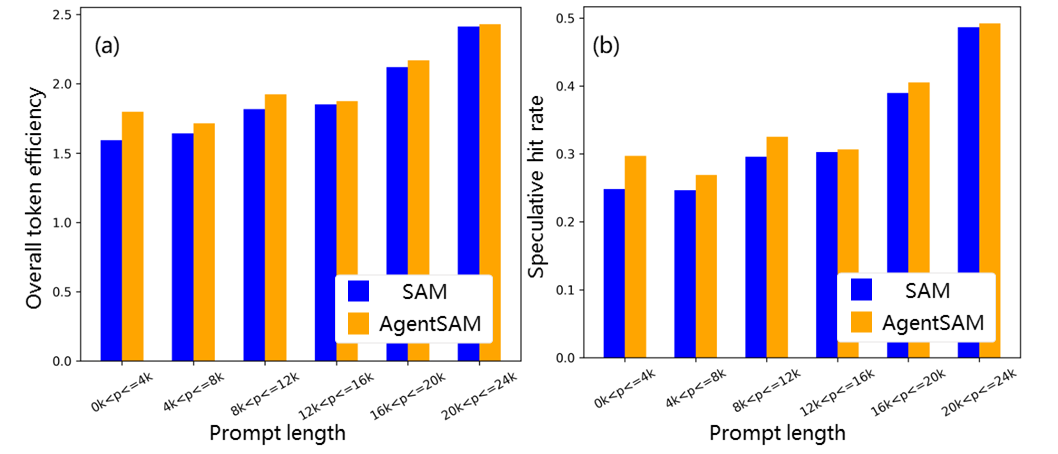}
    \caption{OTE and SHR of SAM and AgentSAM evaluated from BrowseComp-zh dataset.}
    \label{fig:placeholder}
\end{figure}

\textbf{Effectiveness of Memory.}
OTE and SHR results consistently show that as the length of the model input increases, both the speculative hit rate and Overall Token Efficiency improve—an outcome that aligns with our expectation that longer contexts provide more repeated sub-sequences for the suffix automaton to exploit. As AgentSAM integrates session-level memory and cross-session memory into the SAM corpus, both metrics are further improved: the memory supplies additional matching subsequences that the SAM can retrieve as speculative tokens, raising hit rates and reducing the number of tokens that must be verified by full decoding. These improvements translate into measurable gains in OTE.

\begin{table}[htbp]
\centering
\caption{TTFT and TPOT with AgentSAM on Ascend }
\label{tab:asynchronous_sam_construction}
\begin{tabular}{lcc}
\toprule
\textbf{Model Configuration} & \textbf{TTFT (ms)} & \textbf{TPOT (ms)} \\

\midrule

Baseline (No speculative decoding) & 347 & 15.5  \\
Synchronous SAM construction & 1752 & 8.5 \\
Asynchronous SAM construction & 364 & 10.4 \\

\midrule
\end{tabular}
\end{table}

\textbf{Effectiveness of Asynchronous SAM Construction.} We evaluated three configurations—no speculation, synchronous SAM construction, and asynchronous (background-thread) SAM construction with a continuously repetitive prompt of 30k tokens, measuring Time-to-First-Token (TTFT) and per-token processing time (TPOT). Synchronous SAM construction consistently increased TTFT by more than 1.0 second relative to the baseline configuration, while asynchronous construction produced TTFT values statistically indistinguishable from the baseline, confirming that moving SAM building into a background thread removes the blocking impact on initial latency. For decoding-stage throughput, the synchronous construction yielded the lowest TPOT ($\sim$8 ms), the asynchronous construction was slightly higher ($\sim$10 ms), and the baseline case was highest ($\sim$15 ms). The reason is straightforward: synchronous speculation is available from the first speculative step, minimizing per-token cost; asynchronous construction, however, incurs a mixture of non-speculative and speculative steps until the SAM is ready, so its observed TPOT lies between synchronous and no-speculation. Taken together, these results indicate that background-thread SAM construction provides the best practical trade-off for agent deployments: it preserves first-token latency while still delivering most of the per-token performance benefit of speculative decoding.



\subsection{Evaluation on the Entire System}

\begin{table}[!ht]
  \centering
  \caption{Overall QPS (Queries Per Second) improvement of different optimization stages under different parallel agent sessions.}
  \label{tab:agent_optimizations_latency}
  \begin{tabular}{lcc}
    \toprule
    \multirow{2}{*}{Method} & \multicolumn{2}{c}{Overall QPS Improvement} \\
    \cmidrule(lr){2-3}
                            & $N_{parallel}=4$ & $N_{parallel}=16$ \\
    \midrule
    Baseline                                            &  1.00$\times$  &  1.00$\times$  \\
    + AgentCollab                                       &  1.32$\times$  &  1.52$\times$  \\
    + AgentCollab + AgentCompress                       &  1.57$\times$  &  2.01$\times$  \\
    + AgentCollab + AgentCompress + AgentSched          &  1.71$\times$  &  2.25$\times$  \\
    + AgentCollab + AgentCompress + AgentSched + AgentSAM
    &  1.97$\times$  &  2.52$\times$  \\
    \bottomrule
  \end{tabular}
\end{table}

We integrate all proposed components into a single serving stack and evaluate its end-to-end QPS using DeepDiverV2-38B deployed with TP8, with the 7B variant deployed with TP2 as the collaborative model in AgentCollab. As shown in Table~\ref{tab:agent_optimizations_latency}, each module brings consistent and cumulative gains over the baseline.

When $N_{\text{parallel}}=4$, introducing AgentCollab alone improves overall QPS by $1.32\times$, and adding AgentCompress and AgentSched further boosts the speedup to $1.57\times$ and $1.71\times$, respectively. With all four components enabled (including AgentSAM), the system achieves a $1.97\times$ improvement. Under higher concurrency ($N_{\text{parallel}}=16$), the benefits are amplified: QPS improves from $1.52\times$ with only AgentCollab to $2.01\times$ with AgentCompress, $2.25\times$ with AgentSched, and finally $2.52\times$ when AgentSAM is also enabled. The marginal speedup of AgentSAM is reduced at high concurrency because speculative decoding is often deactivated for very large batches to avoid slowing down overall inference.

These results highlight that our optimizations are complementary at the system level: collaboration reduces redundant computation across agents, compression and scheduling improve cache and GPU utilization under load, and speculative decoding further accelerates token generation for long-tail tasks. Overall, the full-stack design more than halves the end-to-end latency in realistic agent applications, demonstrating that careful, system-wide co-design is crucial for scaling multi-agent LLM applications in practice.
\section{Conclusion and Future Works}

The pursuit of efficient autonomous agents requires a fundamental shift in perspective: from micro-optimizing isolated components to macro-optimizing the end-to-end task workflow. Our findings demonstrate that na\"ive acceleration can be a double-edged sword, often introducing overheads that negate potential gains. A successful strategy must be holistic and synergistic, intelligently orchestrating architectural optimizations like AgentCollab and AgentCompress with inference-level strategies like AgentSched and AgentSAM. By adopting this task-centric view, AgentInfer successfully reduces ineffective token consumption by over 50\% and achieves 1.8x-2.5x speedup on end-to-end latency, proving that efficiency need not come at the cost of cognitive capability. These results pave the way for a future of truly fast, reliable, and scalable autonomous agent systems that are optimized not just for speed, but for the successful completion of complex, long-horizon tasks.


\section*{Acknowledgments}

We would like to thank our Project Members for their valuable contributions and support throughout this project:

\textbf{Contributors:}
Jie Chen, Jiahong Zhang, Yijun Hong, Fang Guo

\textbf{Project Sponsors:}
Mingming Zhu, Yaoyuan Wang, Zhenhua Dong, Peifeng Qin, Baochuan Yang, Yunhe Wang

\bibliography{iclr2026_conference}
\bibliographystyle{IEEEtran}

\end{document}